\documentclass[twoside,11pt]{article}

\usepackage[utf8]{inputenc} % allow utf-8 input
\usepackage[T1]{fontenc}    % use 8-bit T1 fonts

\usepackage{jmlr2e}

\usepackage{blindtext}
\usepackage{setspace}

\usepackage[]{algorithm2e}
\usepackage{listings}
\usepackage{xcolor}
\definecolor{codebg}{rgb}{0.95,0.95,0.97}
\definecolor{codecomment}{rgb}{0.5,0.5,0.5}
\definecolor{codekeyword}{rgb}{0.8,0.25,0.2}
\definecolor{codename}{rgb}{0.0,0.3,0.9}
\definecolor{codeidentifier}{rgb}{0.15,0.15,0.15}

\usepackage{url}            % simple URL typesetting
\usepackage{booktabs}       % professional-quality tables
\usepackage{amsfonts}       % blackboard math symbols
\usepackage{nicefrac}       % compact symbols for 1/2, etc.
\usepackage{microtype}      % microtypography

\usepackage{amsmath}

\ShortHeadings{EagerPy}{Rauber, Bethge, and Brendel}
\firstpageno{1}

\newcommand{\appendixref}[1]{\hyperref[#1]{Appendix~\ref{#1}}}

\begin{document}

\lstset{%
	backgroundcolor=\color{codebg},
	keywordstyle=\color{black}\textbf,
	%otherkeywords={=,>,*,-,:},
	identifierstyle=\color{codeidentifier},
	commentstyle=\color{codecomment}\itshape,
	%emph={np,square,where,argsort,cumsum,ediff1d,flatnonzero,sqrt},
	emph={clipping_aware_rescaling},
	emphstyle=\color{black},
	basicstyle=\small\ttfamily,
	numbers=left,
	numbersep=10pt,
	numberstyle=\tiny\color{codecomment},
	abovecaptionskip=\medskipamount,
	captionpos=b,
	framexleftmargin=5pt,
	xleftmargin=5pt,
	framextopmargin=1pt,	
	framexbottommargin=1pt,
	frame=lines,
	language=Python}

\title{EagerPy: Writing Code That Works Natively with\\ PyTorch, TensorFlow, JAX, and NumPy}

\author{%
\\
\name \vspace*{3pt}Jonas Rauber$^{1,2}$ \email jonas.rauber@bethgelab.org\\
\name \vspace*{3pt}Matthias Bethge$^{1,3,\dagger}$ \email matthias.bethge@bethgelab.org\\
\name \vspace*{3pt}Wieland Brendel$^{1,3,\dagger}$ \email wieland.brendel@bethgelab.org\\
\AND
\addr $^{1}$ Tübingen AI Center, University of T{\"u}bingen, Germany\\
\addr $^{2}$ International Max Planck Research School for Intelligent Systems, T{\"u}bingen, Germany\\
\addr $^{3}$ Bernstein Center for Computational Neuroscience T{\"u}bingen, Germany\\
\addr $^{\dagger}$ joint senior authors
}

% \editor{Kevin Murphy and Bernhard Sch{\"o}lkopf}
\editor{}

\maketitle

\begin{abstract}%
	EagerPy is a Python framework that lets you write code that automatically works natively with PyTorch, TensorFlow, JAX, and NumPy. Library developers no longer need to choose between supporting just one of these frameworks or reimplementing the library for each framework and dealing with code duplication. Users of such libraries can more easily switch frameworks without being locked in by a specific 3rd party library. Beyond multi-framework support, EagerPy also brings comprehensive type annotations and consistent support for method chaining to any framework. The latest documentation is available online at \url{https://eagerpy.jonasrauber.de} and the code can be found on GitHub at \url{https://github.com/jonasrauber/eagerpy}.
\end{abstract}


\section{Introduction}
\label{sec:introduction}

The recent advances in deep learning go hand in hand with the development of more and more deep learning frameworks. These frameworks provide high-level yet efficient APIs for automatic differentiation and GPU acceleration and make it possible to implement extremely complex and powerful deep learning models with relatively little and simple code.

Originally, many of the popular frameworks like Theano \citep{theano}, Caffe \citep{caffe}, MXNet \citep{mxnet}, TensorFlow \citep{tensorflow}, and CNTK \citep{cntk} used a graph-based approach. The user first defines a static data flow graph that can then be efficiently differentiated, compiled, and executed on GPUs. Knowing the whole computation graph ahead of time is useful for achieving high performance. It can, however, make it difficult to debug models and to implement dynamic models with changing graphs such as RNNs.

More recently, eager execution of deep learning models has become the dominant approach in deep learning research. Instead of building a static data flow graph ahead of time, eager execution frameworks provide a define-by-run API that builds dynamic, temporary graphs on the fly. The first popular implementations of this approach were Torch \citep{torch}, Chainer \citep{chainer}, and DyNet \citep{dynet}. Using the define-by-run approach, they made it much easier to debug models and to implement dynamic computation graphs such as RNNs. Originally, this came at the cost of lower performance or the need to use less popular programming languages. This changed when PyTorch \citep{pytorch} combined the advantages of the different eager execution frameworks, that is it combined high performance---competitive to graph-based frameworks---with an easy-to-use define-by-run \emph{Python} API. With the introduction of TensorFlow Eager \citep{tensorfloweager} and the switch to eager execution in TensorFlow 2, eager execution is now being used by the two dominant deep learning frameworks, PyTorch and TensorFlow.

Despite these similarities between PyTorch and TensorFlow 2, it is not easily possible to write framework-agnostic code that directly works with both frameworks. At the semantic level, the most fundamental difference lies in the APIs for automatic differentiation. In PyTorch, gradients are requested using an in-place \texttt{requires\_grad\_()} call, zeroed out using the \texttt{zero\_grad()} function, backpropagated by calling \texttt{backward()} and finally read using the \texttt{.grad} attribute. TensorFlow offers a more high-level \texttt{GradientTape} context manager to track gradients and a \texttt{tape.gradient} function to query gradients. Beyond that, the APIs of PyTorch and TensorFlow 2 differ a lot at the syntactic level, e.g.\ in how they name parameters (e.g. \texttt{dim} vs. \texttt{axis}), classes (e.g. \texttt{CrossEntropyLoss} vs. \texttt{CategoricalCrossentropy}), and functions (e.g. \texttt{sum} vs. \texttt{reduce\_sum}), and whether they support method chaining.

EagerPy resolves these differences between PyTorch and TensorFlow 2 by providing a single unified API that transparently maps to the different underlying frameworks without computational overhead. This is similar to how Keras \citep{chollet2015keras} unified the graph-based APIs of TensorFlow 1 and Theano. The difference is that EagerPy focuses on eager execution instead of graph building. In addition, EagerPy's approach is very transparent and allows users to easily combine framework-agnostic EagerPy code with framework-specific code. This makes it possible to gradually adopt EagerPy for individual functions.

Supporting additional eager execution frameworks in EagerPy is just a matter of specifying the necessary translations. EagerPy therefore also comes with support for JAX \citep{jax}, a relatively new framework that has recently gotten a lot of traction thanks to its functional design, NumPy-compatible API and innovative features such as automatic vectorization. In fact, EagerPy's approach to unify the different APIs for automatic differentiation borrows a lot from the high-level functional automatic differentiation API in JAX. Finally, EagerPy also supports NumPy \citep{numpy} as yet another backend, though of course NumPy neither supports automatic differentiation nor GPU acceleration.

EagerPy thus makes it possible to write framework-agnostic code that works natively with PyTorch, TensorFlow, JAX, and NumPy. In a first step, developers of new libraries profit from this because they no longer need to choose between supporting just one of these frameworks or reimplementing their library for each framework and dealing with code duplication. In a second step, the users of these libraries profit because they can more easily switch frameworks without being locked in by a specific 3rd party library.

Beyond that, even users of only a single framework can benefit from EagerPy because it brings all of EagerPy's API improvements such as comprehensive type annotations and consistent support for method chaining to each supported framework.

\section{Design \& Implementation}

EagerPy is build with four design goals in mind. The two key goals are to provide a unified API for eager execution (\autoref{sec:unified}) and to maintain the native performance of the underlying frameworks (\autoref{sec:performance}). These two key goals define what EagerPy is and are core to its design. The two additional goals, a fully chainable API (\autoref{sec:chainable}) and comprehensive type checking support (\autoref{sec:type_checking}), make EagerPy easier and safer to work with than the underlying framework-specific APIs. Despite these changes and improvements, we try to not unnecessarily sacrifice familiarity. Whenever it makes sense, the EagerPy API follows the standards set by NumPy, PyTorch, and JAX.

\subsection{Unified API}
\label{sec:unified}

To achieve syntactic consistency, we define an abstract \texttt{Tensor} class with the appropriate methods and an instance variable holding the native tensor, and then implement a specific subclass for each supported framework. For many operations such as \texttt{sum} or \texttt{log} this is as simple as calling the underlying framework, for others it is slightly more work. The most difficult part is unifying the automatic differentiation APIs. PyTorch uses a low-level autograd API that allows but also requires precise control over the backpropagation (see \autoref{sec:introduction} for some details). TensorFlow uses a slightly higher-level API based on gradient tapes. And JAX uses a very high-level API based on differentiating functions. To unify them, EagerPy mimics JAX's high-level functional API and reimplements it in PyTorch and TensorFlow. EagerPy exposes it through its \texttt{value\_and\_grad\_fn()} function (\appendixref{sec:autodiff_code}).

Being able to write code that automatically works with all supported frameworks requires not only syntactic but also semantic unification. To guarantee this, EagerPy comes with a huge test suite that verifies the consistency between the different framework-specific subclasses. It is automatically run on all pull-requests and needs to pass before new code can be merged. The test suite also acts as the ultimate reference for which operations and which parameter combinations are supported. This avoids inconsistencies between documentation and implementation and in practice results in a test-driven development process.

\subsection{Native Performance}
\label{sec:performance}

Without EagerPy, code that wants to interface with different deep learning frameworks has to go through NumPy. This requires expensive memory copies between CPU (NumPy) and GPU (PyTorch, TensorFlow, JAX) and vice versa. Furthermore, many computations are then only executed on CPU. To avoid this, EagerPy just keeps references to the original native framework-specific tensors (e.g.\ the PyTorch tensor on GPU) and delegates all operations to the respective framework. This introduces virtually no computational overhead.

\subsection{Fully Chainable API}
\label{sec:chainable}

Many operations such as \texttt{sum} or \texttt{square} take a tensor and return one. Often, these operations are applied sequentially, e.g.\ \texttt{square}, \texttt{sum}, and \texttt{sqrt} to compute the $L_2$ norm. In EagerPy, all operations are available as methods on the tensor object. This makes it possible to chain the operations in their natural order: \texttt{x.square().sum().sqrt()}. In contrast, NumPy, for example, requires an inverted order of operations: \texttt{np.sqrt(np.square(x).sum())}.

\subsection{Type Checking}
\label{sec:type_checking}

In Python 3.5, the Python syntax was extended to support type annotations \citep{pep484}. Even with type annotations, Python remains a dynamically typed programming language and all type annotations are currently ignored during runtime. They can however be checked by static code analyzers before running the code.

EagerPy comes with comprehensive type annotations of all parameters and return values and checks them using Mypy \citep{mypy}. This helps us catch bugs in EagerPy that would otherwise stay undetected. EagerPy users can further benefit from this by type annotating their own code and thus automatically checking it against EagerPy's function signatures. This is particularly useful because TensorFlow, NumPy, and JAX do not currently provide type annotations themselves.

\section{Examples}
\autoref{alg:main_example} shows a generic EagerPy \texttt{norm} function that can be called with a native tensor from any framework and returns its norm, again as a native tensor from the same framework. More examples with detailed explanations can be found in \appendixref{sec:converting} and \appendixref{sec:generic_functions}.
\begin{lstlisting}[belowskip=0pt,label={alg:main_example},caption={A framework-agnostic \texttt{norm} function}]
import eagerpy as ep

def norm(x):
    x = ep.astensor(x)  # native tensor to EagerPy tensor
    result = x.square().sum().sqrt()
    return result.raw  # EagerPy tensor to native tensor
\end{lstlisting}

\vspace*{-6pt}

\section{Use Cases}

Foolbox \citep{rauber2017foolbox} is a highly popular adversarial attacks library (more than 220 citations and 1.500 stars on GitHub) that has long supported different deep learning frameworks through a common NumPy interface. With Foolbox 3.0 aka Foolbox Native \citep{rauber2020foolboxnative}, it has been completely reimplemented using EagerPy. It now achieves native performance while still supporting different frameworks using a single code base.

While EagerPy was specifically created with Foolbox in mind, it is now being adopted by other libraries as well. GUDHI \citep{gudhi}, for example, is a library for computational topology. It uses EagerPy to support automatic differentiation in PyTorch, TensorFlow, and JAX without code duplication. Moreover, EagerPy also makes it easy to share framework-agnostic reference implementations of algorithms \citep{rauber2020fast}.

\section{Conclusion}

EagerPy provides a unified API to PyTorch, TensorFlow, JAX, and NumPy without sacrificing performance. Automatic tests guarantee consistency across frameworks. Automatic deployments encourage rapid releases. Comprehensive type annotations help detecting bugs early. Consistent support for method chaining enables beautiful code. And being the foundation of the popular Foolbox library ensures continuous development.

\newpage

\acks{J.R. acknowledges support from the Bosch Research Foundation (Stifterverband, grant T113/30057/17) and the International Max Planck Research School for Intelligent Systems (IMPRS-IS). This work was supported by the German Federal Ministry of Education and Research (BMBF): Tübingen AI Center, FKZ: 01IS18039A, and by the Intelligence Advanced Research Projects Activity (IARPA) via Department of Interior/Interior Business Center (DoI/IBC) contract number D16PC00003. The U.S. Government is authorized to reproduce and distribute reprints for Governmental purposes notwithstanding any copyright annotation thereon. Disclaimer: The views and conclusions contained herein are those of the authors and should not be interpreted as necessarily representing the official policies or endorsements, either expressed or implied, of IARPA, DoI/IBC, or the U.S. Government.
}

\bibliography{main}

\newpage

\appendix

\section{Converting Between EagerPy and Native Tensors}
\label{sec:converting}

A native tensor could be a PyTorch GPU or CPU tensor (\autoref{alg:torch_tensor}), a TensorFlow tensor (\autoref{alg:tensorflow_tensor}), a JAX array (\autoref{alg:jax_array}), or a NumPy array (\autoref{alg:numpy_array}).

\begin{lstlisting}[label={alg:torch_tensor},caption={A native PyTorch tensor}]
import torch
x = torch.tensor([1., 2., 3., 4., 5., 6.])
\end{lstlisting}

\begin{lstlisting}[label={alg:tensorflow_tensor},caption={A native TensorFlow tensor}]
import tensorflow as tf
x = tf.constant([1., 2., 3., 4., 5., 6.])
\end{lstlisting}

\begin{lstlisting}[label={alg:jax_array},caption={A native JAX array}]
import jax.numpy as np
x = np.array([1., 2., 3., 4., 5., 6.])
\end{lstlisting}

\begin{lstlisting}[label={alg:numpy_array},caption={A native NumPy array}]
import numpy as np
x = np.array([1., 2., 3., 4., 5., 6.])
\end{lstlisting}

No matter which native tensor you have, it can always be turned into the appropriate EagerPy tensor using \texttt{ep.astensor}. This will automatically wrap the native tensor with the correct EagerPy tensor class. The original native tensor can always be accessed using the \texttt{.raw} attribute. A full example is shown in \autoref{alg:convert}.

\begin{lstlisting}[label={alg:convert},caption={Converting between EagerPy and native tensors}]
# x should be a native tensor (see above)
# for example:
import torch
x = torch.tensor([1., 2., 3., 4., 5., 6.])

# Any native tensor can easily be turned into an EagerPy tensor
import eagerpy as ep
x = ep.astensor(x)

# Now we can perform any EagerPy operation
x = x.square()

# And convert the EagerPy tensor back into a native tensor
x = x.raw
# x will now again be a native tensor (e.g. a PyTorch tensor)
\end{lstlisting}

Especially in functions, it is common to convert all inputs to EagerPy tensors. This could be done using individual calls to \texttt{ep.astensor}, but using \texttt{ep.astensors} this can be written even more compactly (\autoref{alg:astensors}).

\begin{lstlisting}[belowskip=0pt,label={alg:astensors},caption={Converting multiple native tensors at once}]
# x, y should be a native tensors (see above)
# for example:
import torch
x = torch.tensor([1., 2., 3.])
y = torch.tensor([4., 5., 6.])

import eagerpy as ep
x, y = ep.astensors(x, y)  # works for any number of inputs
\end{lstlisting}

\section{Implementing Generic Framework-Agnostic Functions}
\label{sec:generic_functions}

Using the conversion functions shown in \appendixref{sec:converting}, we can already define a simple framework-agnostic function (\autoref{alg:simple_norm}). This function can be called with a native tensor from any framework and it will return the norm of that tensor, again as a native tensor from that framework (\autoref{alg:calling_norm_pytorch}, \autoref{alg:calling_norm_tensorflow}).

\begin{lstlisting}[label={alg:simple_norm},caption={A simple framework-agnostic \texttt{norm} function}]
import eagerpy as ep

def norm(x):
    x = ep.astensor(x)
    result = x.square().sum().sqrt()
    return result.raw
\end{lstlisting}

\begin{lstlisting}[label={alg:calling_norm_pytorch},caption={Calling the \texttt{norm} function using a PyTorch tensor}]
import torch
norm(torch.tensor([1., 2., 3.]))
# tensor(3.7417)
\end{lstlisting}

\begin{lstlisting}[label={alg:calling_norm_tensorflow},caption={Calling the \texttt{norm} function using a TensorFlow tensor}]
import tensorflow as tf
norm(tf.constant([1., 2., 3.]))
# <tf.Tensor: shape=(), dtype=float32, numpy=3.7416575>
\end{lstlisting}

If we would call the function in \autoref{alg:simple_norm} with an EagerPy tensor, the \texttt{ep.astensor} call would simply return its input. The \texttt{result.raw} call in the last line would however still extract the underlying native tensor. Often it is preferably to implement a generic function that not only transparently handles any native tensor but also EagerPy tensors, that is the return type should always match the input type. This is particularly useful in libraries like Foolbox that allow users to work with EagerPy and native tensors. To achieve that, EagerPy comes with two derivatives of the above conversion functions: \texttt{ep.astensor\_} and \texttt{ep.astensors\_}. Unlike their counterparts without an underscore, they return an additional inversion function that restores the input type. If the input to \texttt{astensor\_} is a native tensor, \texttt{restore\_type} will be identical to \texttt{.raw}, but if the original input was an EagerPy tensor, \texttt{restore\_type} will not call \texttt{.raw}. With that, we can write generic framework-agnostic functions that work transparently for any input (\autoref{alg:better_generic}, \autoref{alg:better_generic_multiple}).

\begin{lstlisting}[label={alg:better_generic},caption={An improved framework-agnostic \texttt{norm} function}]
import eagerpy as ep

def norm(x):
    x, restore_type = ep.astensor_(x)
    result = x.square().sum().sqrt()
    return restore_type(result)
\end{lstlisting}

\begin{lstlisting}[belowskip=0pt,label={alg:better_generic_multiple},caption={Converting and restoring multiple inputs using \texttt{ep.astensors\_}}]
import eagerpy as ep

def example(x, y, z):
    (x, y, z), restore_type = ep.astensors_(x, y, z)
    result = (x + y) * z
    return restore_type(result)
\end{lstlisting}

\section{Automatic Differentiation in EagerPy}
\label{sec:autodiff_code}

EagerPy uses a functional approach to automatic differentiation. You first define a function that will then be differentiated with respect to its inputs. This function is then passed to \texttt{ep.value\_and\_grad} to evaluate both the function and its gradient (\autoref{alg:autodiff}). More generally, you can also use \texttt{ep.value\_aux\_and\_grad} if your function has additional auxiliary outputs and \texttt{ep.value\_and\_grad\_fn} if you want the gradient function without immediately evaluating it at some point $x$.

\begin{lstlisting}[label={alg:autodiff},caption={Using \texttt{ep.value\_and\_grad} for automatic differentiation in EagerPy}]
import torch
x = torch.tensor([1., 2., 3.])

# The following code works for any framework, not just Pytorch!

import eagerpy as ep
x = ep.astensor(x)

def loss_fn(x):
    # this function takes and returns an EagerPy tensor
    return x.square().sum()

print(loss_fn(x))
# PyTorchTensor(tensor(14.))

print(ep.value_and_grad(loss_fn, x))
# (PyTorchTensor(tensor(14.)), PyTorchTensor(tensor([2., 4., 6.])))
\end{lstlisting}

\end{document}